\def\cvprPaperID{6072}
\def\confName{ICCV}
\def\confYear{2023}

\def\paperTitle{Tree-Structured Shading Decomposition}

\def\authorBlock{
    Chen Geng\footnotemark[1] \qquad
    Hong-Xing Yu\footnotemark[1] \qquad
    Sharon Zhang \qquad
    Maneesh Agrawala \qquad
    Jiajun Wu \\[0.5em]
    Stanford University \\
}

\newif\ifreview 
\newif\ifarxiv \newcommand{\arxiv}{\arxivtrue}
\newif\ifcamera 
\newif\ifrebuttal 

\arxiv %

\pdfoutput=1
\documentclass[10pt,twocolumn,letterpaper]{article}
\ifreview \usepackage[review]{cvpr} \fi
\ifarxiv \usepackage[pagenumbers]{cvpr} \fi
\ifrebuttal \usepackage[rebuttal]{cvpr} \fi
\ifcamera \usepackage{cvpr} \fi

\usepackage[accsupp]{axessibility}
\usepackage{graphicx}
\usepackage{amsmath}
\usepackage{amssymb}
\usepackage{booktabs}
\usepackage{float}
\usepackage{mathtools}
\usepackage{amsmath,array,graphicx}
\usepackage{kantlipsum}
\usepackage{inconsolata}

\usepackage{times}
\usepackage{microtype}
\usepackage{epsfig}
\usepackage[table,xcdraw,dvipsnames]{xcolor}
\usepackage{caption}
\usepackage{float}
\usepackage{placeins}
\usepackage{color, colortbl}
\usepackage{stfloats}
\usepackage{enumitem}
\usepackage{tabularx}
\usepackage{xstring}
\usepackage{multirow}
\usepackage{xspace}
\usepackage{url}
\usepackage{subcaption}
\usepackage[hang,flushmargin]{footmisc}
\newcommand{\myparagraph}[1]{\vspace{0.1cm}\noindent\textbf{#1}}
\ifcamera \usepackage[accsupp]{axessibility} \fi

\ifarxiv  \fi

\newcommand{\R}[1]{{%
    \textbf{%
        \ifstrequal{#1}{1}{\textcolor{red}{R#1}}{%
        \ifstrequal{#1}{2}{\textcolor{blue}{R#1}}{%
        \ifstrequal{#1}{3}{\textcolor{magenta}{R#1}}{%
        \ifstrequal{#1}{4}{\textcolor{teal}{R#1}}{%
                           \textcolor{cyan}{R#1}%
        }}}}%
    }%
}}

\usepackage{xr-hyper}

\makeatletter
\newcommand*{\addFileDependency}[1]{
  \typeout{(#1)}
  \@addtofilelist{#1}
  \IfFileExists{#1}{}{\typeout{No file #1.}}
}

\makeatother

\usepackage[pagebackref,breaklinks,colorlinks]{hyperref}
\usepackage[capitalize]{cleveref}
\crefname{section}{Sec.}{Secs.}
\crefname{table}{Table}{Tables}
\crefname{figure}{Fig.}{Figs.}

\begin{document}
\title{\paperTitle}
\author{\authorBlock}
\twocolumn[\maketitle
\begin{center}
    \includegraphics[width=0.95\linewidth]{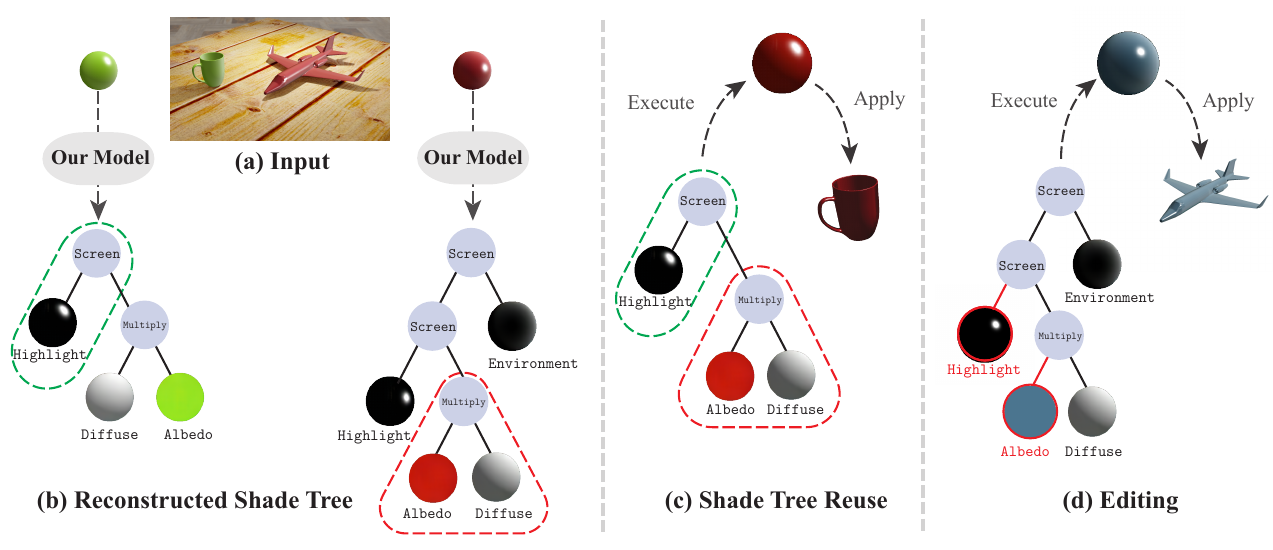}
    \end{center} \vspace{-1.5em}
    \captionof{figure}{\textbf{Decomposing shading into a tree-structured representation.} (a) Our method enables the decomposition of given shading into a (b) \textit{shade tree}. (c) This representation can be reused to generate new \textit{shade trees} and (d) edit the shading of objects. 
    }

    \label{fig:teaser}\bigbreak]
\let\thefootnote\relax\footnotetext{$^*$Equal contribution.}
\begin{abstract}
We study inferring a tree-structured representation from a single image for object shading. Prior work typically uses the parametric or measured representation to model shading, which is neither interpretable nor easily editable. We propose using the \textit{shade tree} representation, which combines basic shading nodes and compositing methods to factorize object surface shading. 
The shade tree representation enables novice users who are unfamiliar with the physical shading process to edit object shading in an efficient and intuitive manner. 
A main challenge in inferring the shade tree is that the inference problem involves both the discrete tree structure and the continuous parameters of the tree nodes. We propose a hybrid approach to address this issue. We introduce an auto-regressive inference model to generate a rough estimation of the tree structure and node parameters, and then we fine-tune the inferred shade tree through an optimization algorithm. We show experiments on synthetic images, captured reflectance, real images, and non-realistic vector drawings, allowing downstream applications such as material editing, vectorized shading, and relighting. Project website: \href{https://chen-geng.com/inv-shade-trees}{https://chen-geng.com/inv-shade-trees}.
\end{abstract}
\section{Introduction}
\label{sec:intro}

Analyzing the shading process in images is fundamental to computer vision and graphics. In particular, the shading process models how the appearances of surfaces are generated from an object's material properties and lighting conditions. Traditional methods formulate it as the problem of intrinsic decomposition, which expresses the shading as the product of reflectance and albedo~\cite{bell2014intrinsic,garces2022survey}. However, this representation is limited in applicability as it assumes a Lambertian surface. Another popular line of works on inverse rendering aims at reconstructing analytical representations~\cite{li2018learning, zhang2021physg, zhang2022iron, munkberg2022extracting, luan2021unified} or measured representations~\cite{lawrence2006inverse} for materials and lighting. Yet, such physical representations are often difficult to interpret in human perception and not user-friendly for image manipulation tasks.

\begin{figure*}[t]
    \centering
    \includegraphics[width=0.95\linewidth]{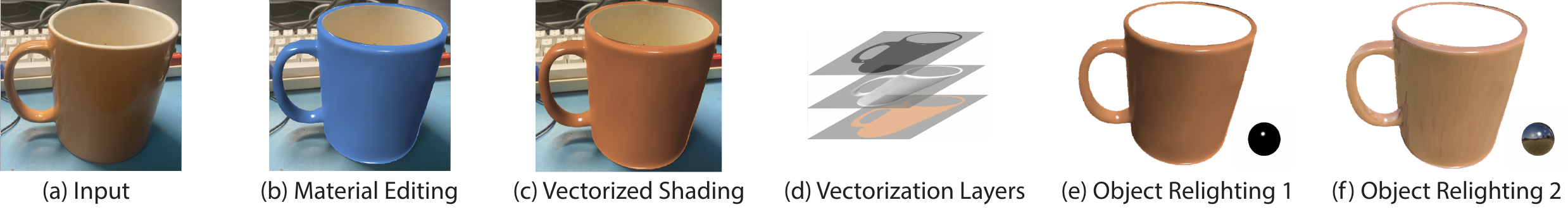}
    \caption{Illustration of downstream applications using the shade tree representation extracted from a single image. For object relighting in (e) and (f), insets show the changed lighting condition.}
    \label{fig:app}
    \vspace{-2mm}
\end{figure*}

The choice of shading representation in inverse graphics is important in that it affects what downstream tasks can be accomplished with that representation. 
The \textit{Shade tree} model is a popular representation for shading in the forward rendering community~\cite{cook1984shade}. 
One important application of this representation is that it models how vector graphics are shaded~\cite{lopez2013depicting}. 
Due to its tree structure, this representation is highly interpretable and easily editable. Thus, it is a worthy and interesting task to recover such representation from visual observations. 

In this work, we study recovering the shade tree representation from a single image. We define our shade tree as a binary tree that contains predefined base nodes (like ``highlight" and ``albedo") and operations (like ``screen mode" and ``mix"). Fig. \ref{fig:teaser} shows examples of our extracted shade trees and subsequently edited materials produced from these extracted trees. In particular, we focus on decomposing the ``shading'' of objects. The input shading can be considered as spherical reflectance maps or ``MatCaps'' obtained from existing pipeline\cite{sloan2001lit,rematas2016deep}.

Despite its desirable high interpretability and editability, inferring such a structured representation from a single image has inherent challenges. First, a shade tree contains both continuous parameters for leaf nodes as well as a discrete tree structure, making it difficult to optimize directly. Second, different combinations of base nodes and operations can lead to equivalent structures, introducing additional ambiguities for deterministic inference methods.
To infer both discrete structure and continuous parameters, we propose a novel two-stage approach to iteratively decompose an input observation into a shade tree. In the first stage, we use an auto-regressive model to recursively decompose nodes to generate an initial tree structure. Then, we perform sub-structure searching and parameter optimization to fine-tune the tree representation. 
To deal with the structural ambiguity, we propose a multiple sampling strategy to allow non-deterministic inference that accounts for the multi-modal distribution of plausible shade trees.

Our extensive experiments show the effectiveness of the proposed approach in decomposing the shading of objects using the shade tree representation. Further, we apply our approach to real shadings and non-realistic vector drawings and demonstrate applications on real images. We demonstrate various downstream tasks in \cref{fig:app}.

In summary, our contributions are three-fold. First, we formulate the problem of inferring \textit{shade trees} from a single image, aiming at understanding the shading of objects with an interpretable representation. Second, we design a novel hybrid approach, integrating an amortized-inference pipeline and an optimization-based solver. Third, we conduct extensive experiments to show the effectiveness of our method and demonstrate potential applications of our method.

\section{Related Work}
\label{sec:related}

\myparagraph{Shade Tree Representation.} The history of using shade trees as a rendering representation in computer graphics can be dated back to the 1980s. Cook \textit{et al.}~\cite{cook1984shade} first proposed this representation in 1984, and subsequent use this representation to model the shading of vector graphics \cite{lopez2013depicting}. 3D software like Blender\cite{blender} uses \textit{node graphs}, a representation similar to \textit{shade trees}.

Few pieces of literature study the problem of inverting such structures. Both Favreau~\etal~\cite{photo2clipart} and Richardt~\etal~\cite{10.1111/cgf.12408} present algorithms to decompose vector graphics into gradient layers, but they do not organize them into tree structures. Lawrence~\etal~\cite{lawrence2006inverse} study the problem of inverting the parameter of leaf nodes given some fixed shade tree structure.  However, our work focuses not only on predicting the parameter of leaf nodes but also on reconstructing the structure of the shade tree.

\myparagraph{Shade Trees v.s. Intrinsic Decomposition / BRDFs.}
Our approach is also related to intrinsic decomposition and inverse rendering. 

Intrinsic decomposition methods seek to decompose images into albedo and reflectance in pixel space without further structures~\cite{garces2012intrinsic,rother2011recovering,bell2014intrinsic,janner2017self,Liu_2020,Li_2018}. The shading structure recovered in this work is flexible, rather than predefined rules (albedo $\times$ reflectance), differing from common intrinsic decomposition tasks.

Traditional inverse rendering methods aim at recovering material, geometry, and lighting from images~\cite{barron2014shape,munkberg2022extracting,hasselgren2022shape,li2020inverse,zhang2021physg,zhang2022iron,luan2021unified,boss2021nerd,Jin2023TensoIR}, using predefined analytical material models such as the Disney BRDF~\cite{burley2012physically}. Compared to the parametric BRDFs, the shade tree focuses on a different level of abstraction. While BRDFs model an element of shading, i.e., \emph{reflectance properties of materials}, it does not model other shading elements such as lighting. Our shade tree models the outcome of shading, i.e., the \emph{appearance}. This involves both material and lighting for real images, as well as other artistic effects in cartoon shadings. Inverting the shade tree representation features advantages including flexibility in shading, interpretability to common users, and high editability.

\myparagraph{Inverse Procedural Graphics.} Procedural graphics generates content algorithmically rather than manually. Textures, biological phenomena, and regular structures like buildings and cities are typically generated with procedural models, with a compact set of variables to direct the generation. Inverse procedural graphics seeks to infer parameters or grammar for procedural models describing such structures. This is often done within specific domains, including urban design and layouts \cite{WuYDZW13, Martinovic_2013_CVPR, 10.1145/2366145.2366187, https://doi.org/10.1111/cgf.14475}, L-systems \cite{10.1145/3394105, https://doi.org/10.1111/j.1467-8659.2009.01636.x}, textures \cite{10.1145/3502431, lefebvre2000analysis, 10.1145/1576246.1531360, LAGAE2010312}, forestry \cite{https://doi.org/10.1111/cgf.12282, 10.1145/3502220}, CSG (Constructive Solid Geometry) trees\cite{10.1145/3272127.3275006, jones2022PLAD, Sharma_2018_CVPR, https://doi.org/10.1111/cgf.13504}, and scene representation \cite{nsd, liu2018learning, Mao2019Program, Li2020Perspective}.

Large material datasets \cite{DADDB18} coupled with differentiable material graph frameworks \cite{Shi2020:ToG, 10.1145/3528233.3530733} have made deep learning methods applicable to procedural material modeling. Given a dataset of training images, Shi~\etal~\cite{Shi2020:ToG} can select an initial graph structure and optimize its parameters to match a target material appearance. Alternatively, Hu ~\etal~\cite{https://doi.org/10.1111/cgf.14591} directly utilizes the latent space of a generative model to transfer material appearance. In a similar vein, Henzler ~\etal~\cite{henzler2021neuralmaterial} embed images into a latent space before generating BRDF parameters. Our method is different from them in that we simultaneously reconstruct the discrete tree structure and the continuous parameters, allowing better adaptation ability to unseen real images.
Generative models have also been applied for creating material representations \cite{10.1145/3414685.3417779, https://doi.org/10.48550/arxiv.2206.05649}. More recently, Guerrero~\etal~\cite{guerrero2022matformer} also shows that transformers are suitable for modeling and generating material graphs, which contain many long-range dependencies. In contrast to generation, we focus on reconstruction from an image.

\section{Method}
\label{sec:method}

\begin{figure*}[t]
\centering
  \includegraphics[width=0.8\textwidth]{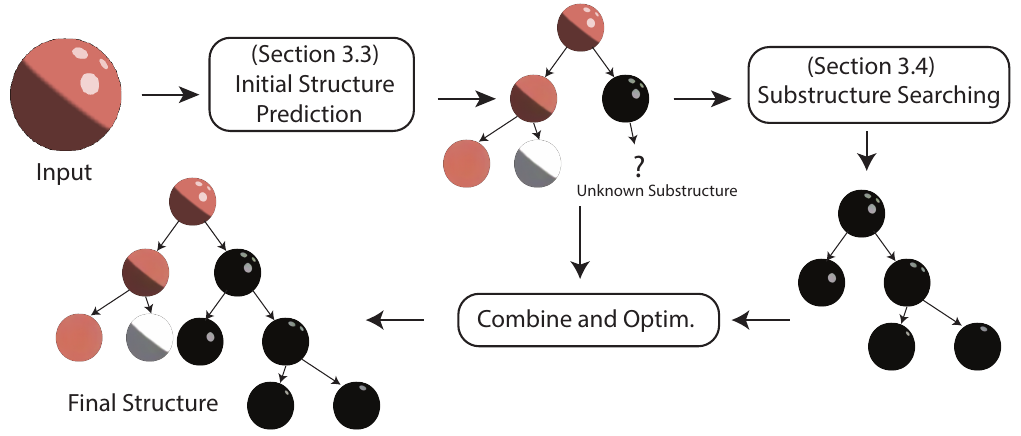}
  \caption{\textbf{The proposed framework for reconstructing shade tree representation.} Our method takes in a shading, and then first does an initial structure prediction in a top-down recursive manner. It is allowed that there is still some unknown substructure from this initial guess. Then the roots of those unknown structures are fed into a substructure searching module, where we perform searching over all possible substructures and optimize the leaf parameter to decide whether this structure is appropriate. After all the substructures are decided, we merge them into the initially predicted tree and get the final structure. We perform an overall optimization on this structure to get the final parameter of the leaf nodes.
  }
  \label{fig:pipeline}
\end{figure*}

We now introduce our tree decomposition pipeline. First, we introduce the context-free grammar used to represent our shade trees (Sec~\ref{subsec:grammar}). Next, we cover the recursive amortized inference used to produce an initial tree structure (Sec~\ref{alg-part-i}). Finally, we explain an additional optimization-based fine-tuning step for decomposing any remaining nodes that were not reliably decomposed by the recursive inference (Sec~\ref{alg-part-ii}).

\subsection{Grammar Specification}\label{subsec:grammar}

\myparagraph{Definition of Shade Trees.}
A \textit{shade tree} is a tree-structured representation for shading. The leaf nodes of the tree structure are all basic shading nodes that cannot be further decomposed. The interior nodes are formulated using a specified composition method taking child nodes as input. By executing the tree structure in a bottom-up manner, we can get complex shading effects.

\myparagraph{Definition of Base Nodes.} We define four basic shading nodes. \texttt{Highlight} nodes represent a single highlight reflected on the surface. \texttt{DiffRef} nodes represent the diffuse reflective component of the material. \texttt{Albedo} nodes are homogeneous nodes with only one uniform color for shading to represent a basic albedo shading. Finally, \texttt{EnvRef} nodes model the specular shading reflecting the surrounding environment.

\myparagraph{Definition of Composition Methods.} We define three compositing methods to construct parent shading nodes from child nodes. The \texttt{Multiply} operator performs a multiplication of its two child nodes. The \texttt{Screen} operation performs a screen mode composition. The \texttt{Mix} operation takes a mask as input and uses the mask to assign different shading nodes to different regions. For \texttt{multiply}, the shading of parent node $p$ is defined as:
\begin{equation}
p = c_l \cdot c_r,
\end{equation}
where $c_l$ and $c_r$ denote the left child and the right child, respectively. The \texttt{screen} operation is given by:
\begin{equation}
p = 1 - (1 - c_l) \cdot (1 - c_r).
\end{equation}
And the \texttt{mix} operation is defined as:
\begin{equation}
    p = m \cdot c_l + (1 - m) \cdot c_r,
\end{equation}
where $m$ denotes a learnable mask.

\myparagraph{Context-Free Grammar.} The definition of the \textit{shade tree} can be formalized to a domain-specific language (DSL) represented by a context-free grammar\cite{hopcroft2001introduction} $G$, as shown in \cref{table:dsl}.

\begin{table}[t]
\centering 
\begin{tabular}{lll}
\toprule
Tree & $\rightarrow$ & \texttt{Mix}(Tree, Tree, Mask)                                       \\
Tree & $\rightarrow$ & \texttt{Multiply}(Tree, Tree)                                           \\
Tree & $\rightarrow$ & \texttt{Screen}(Tree, Tree)                                             \\
Tree & $\rightarrow$ & \texttt{Albedo}(Color=Var) \\
Tree & $\rightarrow$ & \texttt{DiffRef}(Lobe=Var, Ambient=Var) \\
Tree & $\rightarrow$ & EnvRef \\
Tree & $\rightarrow$ & \texttt{Highlight}(Lobe=Var, Sharpness=Var) \\
EnvRef & $\rightarrow$ & a environment map \\
Var  & $\rightarrow$ & free continuous variable \\
Mask & $\rightarrow$ & a map with 0 and 1 \\

\bottomrule
\end{tabular}
\caption{Context-free grammar for the DSL representing \textit{shade tree} structure. More details of the DSL can be found in the supplement. 
}
\label{table:dsl}
\end{table}

\subsection{Overview of Algorithm}

The proposed algorithm contains two stages. We show an overview in \cref{fig:pipeline}. In the first stage, we aim to recover the initial structure of the \textit{shade tree} using a recursive amortized inference decomposition module (\cref{fig:pipeline} top). In the second stage, we decompose the nodes that are not successfully solved in the first stage and recover the parameters of leaf nodes using an evolution-based optimization algorithm (\cref{fig:pipeline} bottom).

The motivation for this two-stage design for decomposition is that we wish to take advantage of the distinct behaviors of these two types of algorithms. The first stage is top-down amortized inference and performs the decomposition layer-by-layer. This approach learns prior knowledge from large-scale training data. Thus, the decomposition is fast but occasionally fails in some corner cases due to the lack of enough capacity to generalize, which is seen as a common problem for learning-based methods. 

Thus, we further introduce the second stage, which employs a classical program synthesis that enumerates all possible structures and does optimization to find the correct solution. Such an enumeration is slower than learning-based methods, yet it has more capacity to generalize to corner cases. By combining these two approaches, our algorithm is effective and efficient in tackling the task.

\subsection{Recursive Amortized Inference}\label{alg-part-i}

\begin{figure*}[t]
\centering
  \includegraphics[width=0.8\textwidth]{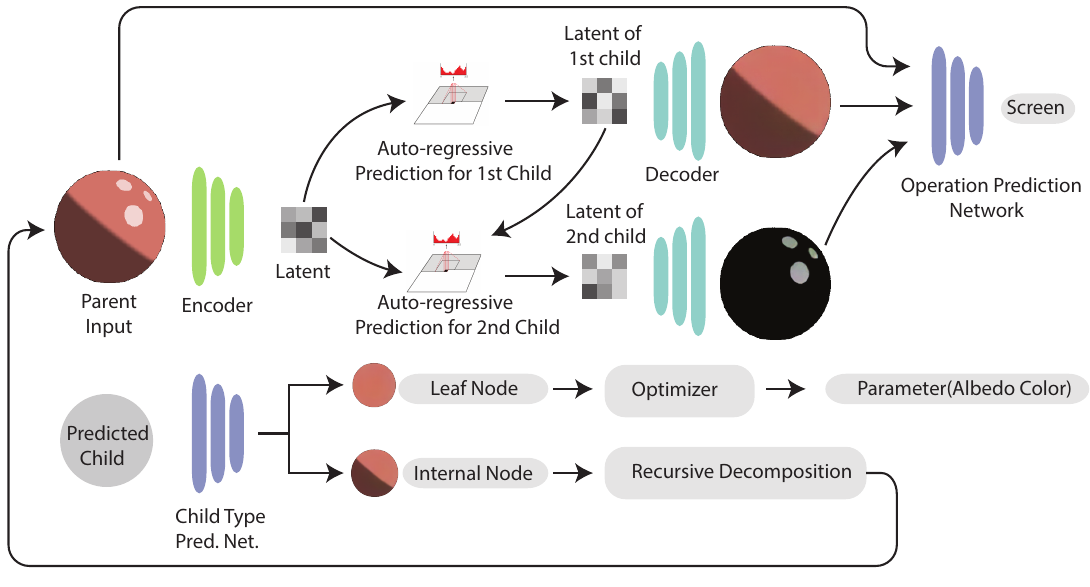}
  \vspace{-0.2cm}
  \caption{\textbf{Illustration of recursive inference module used in the initial structure prediction.} The structure prediction is performed in a top-down recursive manner. In each step, we feed the current parent node into the module, and it is first encoded to a discrete latent code using VQ-VAE. Then the latent code is fed into the first auto-regressive module to predict the latent code for the first child. After that, the latent of the first child and the parent are both fed into the second auto-regressive module to get latent of the second child. The latent codes are decoded into images for children nodes. Afterward, the parent node and the children nodes are fed into the operation prediction network to predict the operation of this step. For each of the predicted children, we use a child-type prediction network to know whether it is a leaf node. If it is not a leaf node, it will be further decomposed. Otherwise, it will be optimized to get its parametric representation.}
  \label{fig:recuri}
\end{figure*}

In the first part, we do the decomposition in a top-down manner recursively and then procedurally generate the entire tree. We maintain a pool of nodes and record their type and linkage for each inference procedure. Initially, there is only one node $\mathbf{I}_0$ in the pool, serving as the root node of the whole tree. At each step, we consider node $\mathbf{I}$ which is neither a leaf nor decomposed. We pass it into our shared single-step component prediction module $M$ and obtain $\{\mathbf{I}_{l}, \mathbf{I}_{r}\} = M(\mathbf{I})$, where $\mathbf{I}_{l}$ and $\mathbf{I}_{r}$ denote the left and right child nodes, respectively. The design of $M$ will be discussed in this section later. The child nodes $\mathbf{I}_{l}$ and $\mathbf{I}_{r}$ are then linked to the parent node $\mathbf{I}$ with new tree edges. All three nodes are then fed into a CNN $f$ which gives
\begin{equation}
\mathbf{p} = f(\mathbf{I}, \mathbf{I}_{l}, \mathbf{I}_{r}),
\end{equation}
where $\mathbf{p}$ is a probability distribution over all  compositing operations in our grammar. The operation with the highest probability is selected as the type of the parent node. 
A separate CNN $g$ that also takes the three nodes as input predicts
\begin{equation}
\Omega = g(\mathbf{I}, \mathbf{I}_{l}, \mathbf{I}_{r}),
\end{equation}
where $\Omega$ is the parameter value of the selected operation.

After predicting the operation and corresponding parameters, we then get the reconstructed parent node $\hat{\mathbf{I}}$ by choosing the correct operation from operation set $S = \{\texttt{mix}, \texttt{screen},\texttt{multiply}\}$ and then get the single-step reconstruction error $\mathcal{L}_\text{recon}$:
\begin{align}
\mathcal{L}_\text{recon} & = ||\mathbf{I} - \hat{\mathbf{I}}||_2, \\ 
\text{where} \quad \quad
\hat{\mathbf{I}} & = S_{\text{argmax}(\mathbf{p})}(\mathbf{I}_{l}, \mathbf{I}_{r}, \Omega).
\label{eq:recon}
\end{align}

To determine whether the predicted child node should be further decomposed, we pass each of $\mathbf{I}_{l}, \mathbf{I}_{r}$ into a child component prediction neural network $h$ and get the probability $\mathbf{q}$ of its type.
If the child node is a leaf node, then we mark it as solved in the node pool, so it is not further decomposed. The previously described procedure ends whenever no more nodes can be decomposed. 

We then describe how the single-step component prediction module $M$ is implemented. The design of $M$ follows two principles:
\begin{enumerate}
  \item The prediction should not be deterministic, i.e., it should allow different kinds of output for this module. This is because many different structures of trees may describe the same tree.
  \item The prediction of the second child should at least depend on the prediction of the first child and the parent node, and the prediction of the first child should depend on the parent node.
\end{enumerate}

Inspired by these two principles, we design a two-step conditional auto-regressive module for prediction. 

\myparagraph{Auto-regressive Inference.} Auto-regressive inference entails a discretized feature space. Thus, we adopt the Vector Quantized Variational Autoencoder (VQ-VAE)~\cite{vqvae2} as our encoder architecture. The latent feature $\mathbf{v}$ of a shading $\mathbf{I}$ is given by: $\mathbf{v} = E(\mathbf{I})$, where $E$ denotes the VQ-VAE encoder (\cref{fig:recuri} left top). Then we further train two conditional PixelSNAIL~\cite{pixelsnail} models to auto-regressively generate two child nodes (\cref{fig:recuri} middle top). Specifically, for the generation of the first child node, we sample the discrete latent representation $\mathbf{v}_{l}$ from the distribution $p_l(\mathbf{v})$ represented using the auto-regressive model, conditioned on the latent code of the parent node. Similarly, the latent code $\mathbf{v}_r$ of the second child is sampled from the distributions $p_r(\mathbf{v}, \mathbf{v}_l)$ encoded by the auto-regressive model, conditioned on the previous child and the parent node. Finally, the child images are decoded using the decoder $D$ to generate the image representation of child nodes.

We build a synthetic dataset to train the previously mentioned modules. Please refer to the supplementary material for the detail of the training.

\myparagraph{Multiple Sampling.} For each auto-regressive inference, we sample $T$ times to make sure that we make the best decomposition decision in each step. We define a criterion to select from multiple samples. First, we need the reconstruction result that combines two child nodes to be as similar as possible to the parent node, which can be indicated from the $\mathcal{L}_\text{recon}$ as described in Eq.~\ref{eq:recon}. 

Further, we wish the derived tree to be as compact as possible by avoiding useless decomposition. We define $\mathcal{L}_\text{sim}$ that represents the similarity between the parent node and the child nodes, which is defined as
\begin{equation}
\mathcal{L}_\text{sim} = -\log(||\mathbf{I} - \mathbf{I}_{l}||_2 + ||\mathbf{I} - \mathbf{I}_{r}||_2).
\end{equation}

We also define $\mathcal{L}_\text{blank}$ and $\mathcal{L}_\text{white}$ to avoid one child being wholly blank or white, which will result in useless decomposition:
\begin{align}
\mathcal{L}_\text{blank} &= -\log(||E(\mathbf{I}_{l}) - \mathbf{1}||_2 + ||E(\mathbf{I}_{r}) - \mathbf{1}||_2),\\
\mathcal{L}_\text{white} &= -\log(||E(\mathbf{I}_{l})||_2 + ||E(\mathbf{I}_{r})||_2).
\end{align}
The final criterion $\mathcal{L}_\text{select}$ is defined as
\begin{equation}
  \mathcal{L}_\text{select} = \mathcal{L}_\text{recon} + \alpha \mathcal{L}_\text{sim} + \beta \mathcal{L}_\text{blank} + \gamma \mathcal{L}_\text{white},
\end{equation}
where $\alpha$, $\beta$, $\gamma$ are hyper-parameters.

\myparagraph{Early-stop Strategy.} The amortized inference module may not successfully decompose every node, which is why we designed the second stage to further decompose those nodes and finetune the whole tree structure. We send a node to the second stage if $\min_T(\mathcal{L}_\text{select}) < \tau$, where $\tau$ is a threshold.

\subsection{Optimization-based Finetuning}\label{alg-part-ii}

For the nodes that cannot be decomposed in the first stage, we search over all possible sub-structures of these nodes and use an optimizer BasinCMA~\cite{basincma} to find the optimal leaf parameters. The optimization target can be defined as
\begin{equation}
  \min_{\lambda} ||R_{s}(\lambda) - \mathbf{I}||_2 + ||F_{\text{VGG}}(R_{s}(\lambda)) - F_{\text{VGG}}(\mathbf{I})||_2,
  \label{optim}
\end{equation}
where $\lambda$ denotes all trainable parameters, $R_s$ represents the renderer under the searched structure $s$ and $F_\text{VGG}$ represents a pretrained VGG network. 

The search over all possible substructures is performed in the order of depth. If the target loss is already smaller than a predefined threshold $\phi$, we stop searching and assume it to be the final substructure.

\myparagraph{Obtaining parametric representation for leaf nodes.} After finalizing the shade tree's structure, we optimize each leaf node using the same target as described in Eq. \ref{optim} to get the parametric representation for each leaf node.

\myparagraph{Optimizing on the whole tree. }Finally, we perform optimization on the parameter of all leaf nodes using the following target:
\begin{equation}
  \min_{\mu} ||R_{S}(\mu) - \mathbf{I}_0||_2 + ||F_{\text{VGG}}(R_{S}(\mu)) - F_{\text{VGG}}(\mathbf{I}_0)||_2,
\end{equation}
where $S$ stands for the finalized structure of the whole tree and $\mu = [\lambda_0, \lambda_1, \cdots, \lambda_N]$.

\begin{table*}[t]
\centering\small
\begin{tabular}{lccccccccc}
\toprule
& \multicolumn{3}{c}{Realistic}                                                        & \multicolumn{3}{c}{Toon}                                                        & \multicolumn{3}{c}{DRM (real-captured)}                          \\ 
\cmidrule(lr){2-4}\cmidrule(lr){5-7}\cmidrule(lr){8-10}
            & \multicolumn{1}{l}{PSNR$\uparrow$} & \multicolumn{1}{l}{SSIM$\uparrow$} & \multicolumn{1}{l}{LPIPS$\downarrow$} & \multicolumn{1}{l}{PSNR$\uparrow$} & \multicolumn{1}{l}{SSIM$\uparrow$} & \multicolumn{1}{l}{LPIPS$\downarrow$} & PSNR$\uparrow$           & SSIM$\uparrow$           & LPIPS$\downarrow$          \\ 
\midrule
CNN         & 26.50                    & 0.967                    & 0.066                     & 27.18                    & 0.959                    & 0.052                     & 14.41          & 0.857          & 0.164          \\
LSTM        & 17.80                    & 0.909                    & 0.154                     & 24.86                    & 0.964                    & 0.067                     & 19.67          & 0.882          & 0.205          \\
Transformer & 18.82                    & 0.930                    & 0.153                     & 26.73                    & 0.971                    & 0.042                     & 17.89          & 0.876          & 0.186      \\
Ours        & \textbf{30.89}           & \textbf{0.974}           & \textbf{0.052}            & \textbf{30.08}           & \textbf{0.972}           & \textbf{0.032}            & \textbf{25.47} & \textbf{0.927} & \textbf{0.150} \\
\bottomrule
\end{tabular}
\caption{\textbf{Quantitative comparison.} Our method greatly surpasses other baselines in all three datasets, benefiting from our design to accurately predict structure and parameters. ``Realistic'' and ``Toon'' are two synthetic datasets and ``DRM'' is a real-captured dataset.}
\label{table:quant}
\vspace{-0.2cm}
\end{table*}

\begin{figure*}[t]
\centering
  \includegraphics[width=0.95\linewidth]{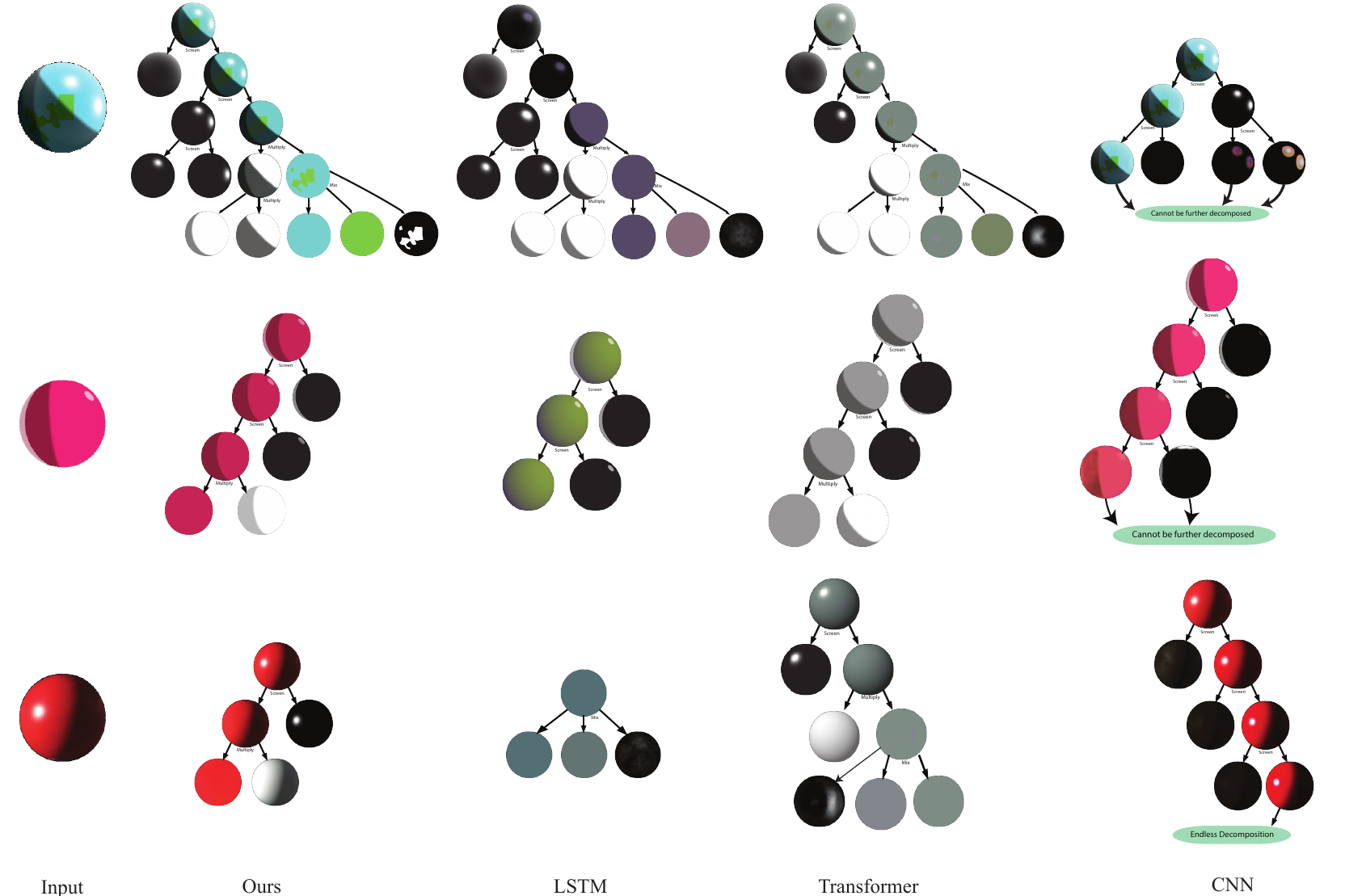}
  \caption{\textbf{Qualitative comparison of different methods on reconstruction.} [\textit{Top row}] shows a sample from \textit{Realistic} dataset, [\textit{Mid row}] shows a sample from \textit{Toon} dataset, and [\textit{Bottom row}] shows a real test sample from the \textit{DRM} dataset \cite{rematas2016deep}. 
  }
  \label{fig:compare}
\end{figure*}
\section{Experiments}
\label{sec:exp}

In this section, we first show our results on shade tree decomposition on a diverse set of example including realistic synthetic images, cartoon-style images, and real images. Then we introduce a visual shading editing analogy experiment which is designed to quantitatively evaluate the decomposition of reconstructed structures. Finally we analyze our method by ablation study results.

\subsection{Results on Decomposition}
\label{sec:decomp}
To evaluate the effectiveness of the proposed method, we conduct a quantitative evaluation of our methods and other baselines on several datasets.

\myparagraph{Datasets.} We evaluate our method on both synthetic and real-captured datasets.

For the synthetic dataset, we generate two styles of datasets, ``Realistic'' and ``Toon'', to show the robustness and broad applicability of the proposed method. For the ``Realistic'' dataset, all the base nodes are represented in a photo-realistic way, imitating how the shading in real life behaves. For the ``Toon'' dataset, we take inspiration from non-photorealistic shading~\cite{gooch2001non} and generate many cartoon-style shading nodes. After generating all base nodes and operation nodes, they are split into two sets, one for training sets and the other for the generation of test sets. Afterward, we apply a recursive algorithm to generate the training and test sets using the specified context-free grammar. The details of the dataset can be found in the supplementary material.

Besides the synthetic datasets, we use the real-captured dataset ``DRM'' collected by Rematas et al. \cite{rematas2016deep} to evaluate the real-world generalizability of the proposed method.

\myparagraph{Baselines.} 
No previous work has tackled a task setting similar to ours. Therefore, we drew inspiration from previous research on grammar decomposition and adapted three competitive baseline frameworks that are widely used in the neural program synthesis and structure induction community for our purpose. 

Our \textbf{CNN} baseline, which utilizes a similar architecture to that in Rim-net \cite{niu2022rim}, employs an encoder-decoder structure to perform single-step decomposition recursively, similar to the first stage of our approach. We also introduced an \textbf{LSTM} baseline, similar to Shape Programs \cite{tian2019learning}, which first uses an encoder to get the latent representation of images and then uses LSTM to predict a sequence of tokens that are subsequently compiled to the shade tree structure. Similarly, our \textbf{Transformer} baseline also predicts the sequence of tokens but adapts a GPT architecture, following Matformer\cite{guerrero2022matformer}. Please note that although the baselines share a similar backbone design with previous literature, they differ due to the different problem settings. The supplementary material contains further details and implementation of the baselines.

\begin{figure*}[t]
\centering
  \includegraphics[width=0.8\linewidth]{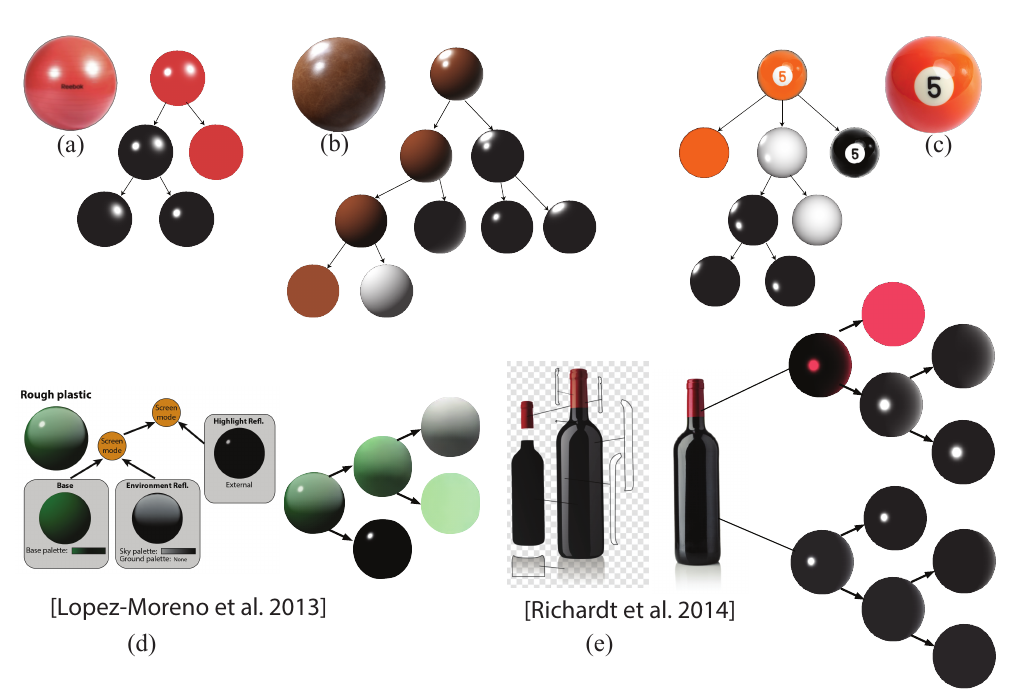}
  \caption{\textbf{Decomposition of in-the-wild real images using our method.} Our method can not only work on synthetic data but can also be widely used in the decomposition of in-the-wild shading. The shadings in (a, b, c) are collected from the Internet. In (d), we show that our method can do decomposition of the shadings from  Lopez-Moreno et al.\cite{lopez2013depicting}. In (e), we compare our method with Richardt et al.\cite{10.1111/cgf.12408}. We first extract shading from the vector drawing, and then we use our method to do decomposition to the shading sphere.}
  \label{fig:more}
\end{figure*}

\begin{figure}[t]
  \centering
\includegraphics[width=0.9\linewidth]{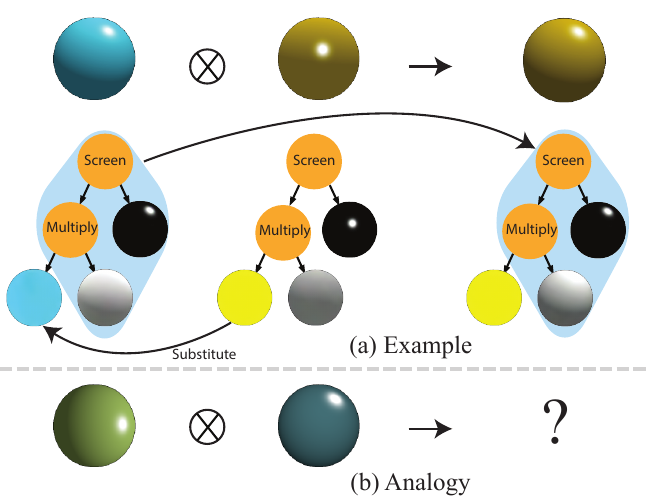}
\caption{\textbf{Illustration of ``Visual Shading Editing Analogy'' that allows quantitative evaluation.} Given an example of shading editing, we wish the same operation could be applied to novel test pairs. For instance, in this case, the example shows us that the edited shading is formed by replacing the albedo node with the albedo node from the second shading ball. Thus, this operation should also be applied to the pair in the bottom row, resulting in a shading ball with a blue base color and an upward highlight.}
\label{fig:ill}
\end{figure}

\myparagraph{Results.} We show the results of this experiment in \cref{fig:compare} and \cref{table:quant}. Our method has the best-reconstructed tree structure among all three methods. The LSTM baseline can predict similar structures to ours; however, it performs poorly in predicting the parameter of leaf nodes, resulting in bad reconstruction results. The CNN baseline predicts in a top-down manner; however, it suffers from ambiguity in the grammar. Thus, it cannot learn a correct mapping between layers, resulting in nodes that cannot be further decomposed or endlessly decomposed in a trivial way. 

Our method can also be applied to the real-world dataset ``DRM'' with satisfactory performance, which can be witnessed from the third column of \cref{table:quant} and the third row of \cref{fig:compare}. The result shows the generalizability and the real-world applicability of the proposed approach.

We also perform decomposition using our model on some in-the-wild internet photos, shown in \cref{fig:more}.

\subsection{Visual Shading Editing Analogy}

\begin{table}[t]
\centering
\small
\begin{tabular}{cccc}
\toprule
      & \multicolumn{1}{c}{PSNR$\uparrow$} & \multicolumn{1}{c}{SSIM$\uparrow$} & \multicolumn{1}{c}{LPIPS$\downarrow$} \\ 
      \midrule
CNN  &    4.79                                &  0.143                                &     0.608                              \\
LSTM &   4.47                                &    0.113                              & 0.547                               \\
Transformer & 5.02 & 0.186 & 0.547\\
Ours  &  \textbf{32.17}                                 &\textbf{0.913}                                 & \textbf{0.078}                                  \\
\bottomrule
\end{tabular}
\caption{\textbf{Quantitative comparison of different methods on the task visual shading editing analogy.} Our method performs the best among all three methods by understanding the tree structure well. The other two methods cannot deal with this task because of their poor decomposition. }
\label{tab:an}
\end{table}

To allow quantitatively evaluating the reconstructed tree structure, we design a task called ``Visual Shading Editing Analogy'' which reflects how well the decomposition is. As illustrated in \cref{fig:ill}, given an input pair of shading, the algorithm should give a hybrid shade tree composed of different subtrees from different nodes, according to the rule shown in the example shading ball pair. 

We generate a dataset containing different types of shading editing to evaluate the performance of different methods on this task. We adopt the same baseline setting in Section \ref{sec:decomp}, introducing the CNN, LSTM, and Transformer baselines. Then we use such methods to decompose given pairs to get their tree-structured representation. The details of the algorithm for making such a visual analogy are described in the supplementary material.

\cref{tab:an} shows the results. Our method surpasses other methods greatly due to a better understanding of the semantic meaning of tree structure. 

\begin{figure}[t]
    \centering
  \includegraphics[width=\linewidth]{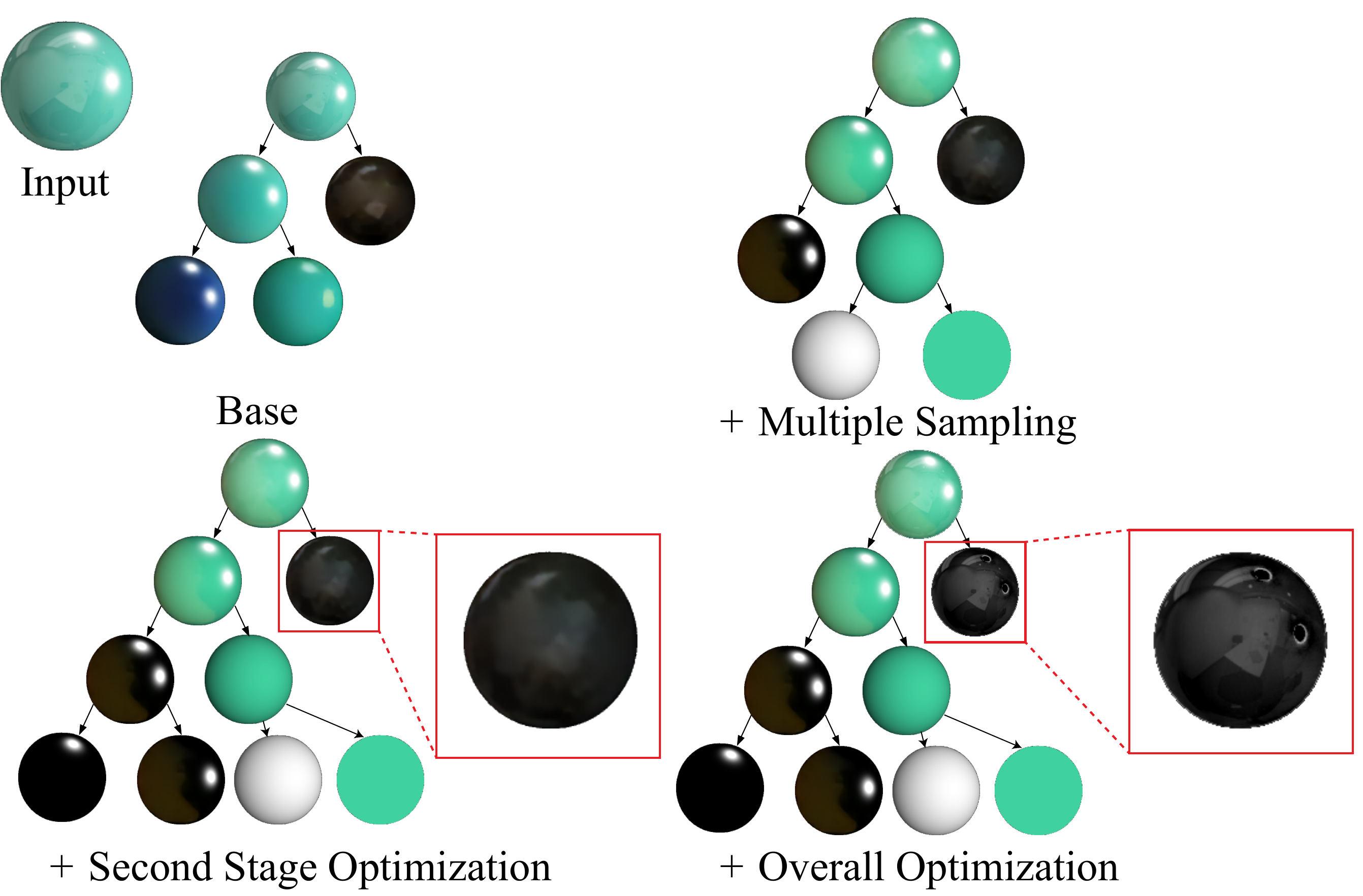}
  \caption{\textbf{Ablation of components.} ``Base'' denotes our method with only 1 sample during inference, ``+Multiple Sampling'' denotes only using the 1st stage with multiple sampling. ``+Second Stage'' denotes using both the first stage and the second stage of the proposed method but does not perform the overall optimization. ``+Overall Optimization'' denotes the full proposed method.}
  \label{fig:ablation}
\end{figure}

\begin{figure}[t!]
\centering
  \vspace{-0.1in}
  \includegraphics[width=0.95\linewidth]{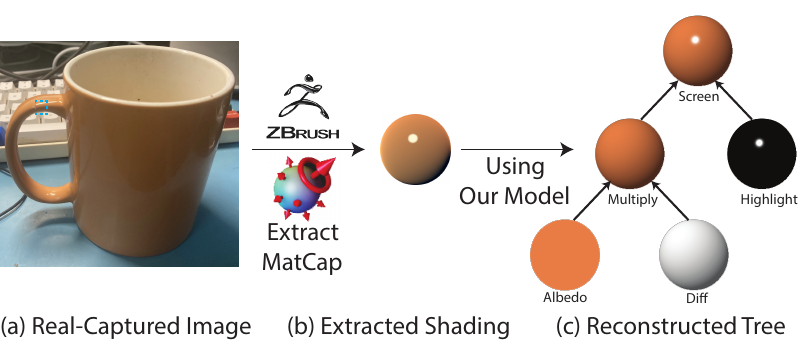}
  \caption{\textbf{Results on a real image of a non-sphere object.} Although the main focus of this work is decomposing the reflectance spheres, it can be applied to non-sphere geometry as well if we use existing tools to first extract sphere reflectance from images first. }
  \label{fig:realdemo}
\end{figure}

\subsection{Ablation Studies} 

To verify the influence of the special design in the proposed pipeline, we do ablation studies on the following three components: multiple sampling, second-stage optimization, and overall optimization, because they are typically non-trivial in previous literature.

\myparagraph{Influence of Multiple Sampling} From \cref{fig:ablation}, it can be observed that multiple sampling improves the result during the 1st stage inference because it can produce several solutions and use the metrics to choose the best one. Without doing this, our model may directly predict a ``reasonable'' one, but not the ``best'' one.

\myparagraph{Influence of Optimization} The second stage of optimization help us to decompose those nodes that are hard to deal with using only the amortized-inference module. By removing such a component, our method cannot decompose the bottom node shown in Fig. \ref{fig:ablation} and thus gives worse results than the full method.

\myparagraph{Influence of Overall Optimization} By introducing the overall optimization at the end of the second stage, our method can further finetune the structure, like giving a better environment reflection.

\subsection{Application on Real-world Images}

Our work focuses on the decomposition of MatCaps or Reflectance Maps\cite{sloan2001lit}. However, the work can be applied to real-world images by using existing tools to first extract the sphere reflectance. In \cref{fig:realdemo}, we show an example of using the proposed method together with an existing tool \textit{ZBrush} to decompose the shading of a real-world capture.

\section{Conclusion}
\label{sec:conclusion}

We have presented a novel method that can effectively and efficiently decompose shading into a tree-structured representation, which enables understanding and editing of the shading in an interpretable way. The first stage of the proposed method uses a pretrained auto-regressive model to predict the structure and parameters of the tree structure. The second stage of the pipeline leverages the parametric representation of each base node and structure searching to find the optimal structure for all nodes that cannot be effectively decomposed in the first stage. The combination of two stages leads to our state-of-the-art performance on several datasets compared to the baselines.

\paragraph{Acknowledgments.} This work was in part supported by Ford, NSF RI \#2211258, AFOSR YIP FA9550-23-1-0127, the Toyota Research Institute (TRI), the Stanford Institute for Human-Centered AI (HAI), Amazon, and the Brown Institute for Media Innovation.

{\small
\bibliographystyle{ieee_fullname}
\bibliography{11_references}
}

\end{document}